\documentclass[10pt,twocolumn,letterpaper]{article}

\usepackage{cvpr}              

\usepackage{graphicx}
\usepackage{amsmath}
\usepackage{amssymb}
\usepackage{booktabs}

\usepackage{times}
\usepackage{epsfig}
\usepackage{tabu}
\usepackage{multirow}
\usepackage{amsfonts}       
\usepackage{xpunctuate}
\usepackage{pifont}
\usepackage{color,colortbl}
\usepackage[rgb,dvipsnames]{xcolor}
\usepackage{subfig}
\usepackage{tabulary,overpic}
\usepackage{booktabs}       
\usepackage[normalem]{ulem}

\usepackage{scalerel}
\usepackage[accsupp]{axessibility}

\newcommand{\cmark}{\ding{51}}%
\newcommand{\xmark}{\ding{55}}%

\definecolor{fyxcolor}{RGB}{0,128,255}
\definecolor{demphcolor}{RGB}{100,100,100}
\definecolor{citecolor}{RGB}{0,0,192} 
\definecolor{GrayBG}{gray}{0.95}

\def\x{$\times$}
\newcommand{\demph}[1]{\textcolor{demphcolor}{#1}}
\newcommand{\app}{\raise.17ex\hbox{$\scriptstyle\sim$}}
\newlength\savewidth\newcommand\shline{\noalign{\global\savewidth\arrayrulewidth
  \global\arrayrulewidth 1pt}\hline\noalign{\global\arrayrulewidth\savewidth}}
\newcommand{\tablestyle}[2]{\setlength{\tabcolsep}{#1}\renewcommand{\arraystretch}{#2}\centering\footnotesize}

\newcommand{\ours}{SCVRL}
\iftrue
\newcommand\del{\bgroup\markoverwith{\textcolor{red}{\rule[0.5ex]{2pt}{1.2pt}}}\ULon}
\else
\fi

\usepackage[pagebackref,breaklinks,colorlinks]{hyperref}

\usepackage[capitalize]{cleveref}
\crefname{section}{Sec.}{Secs.}
\Crefname{section}{Section}{Sections}
\Crefname{table}{Table}{Tables}
\crefname{table}{Tab.}{Tabs.}

\def\cvprPaperID{56} 
\def\confName{CVPR L3D-IVU workshop}
\def\confYear{2022}

\usepackage{overpic}
\usepackage{enumitem} 
\usepackage{overpic} 
\usepackage{color}

\definecolor{turquoise}{cmyk}{0.65,0,0.1,0.3}
\definecolor{purple}{rgb}{0.65,0,0.65}
\definecolor{dark_green}{rgb}{0, 0.5, 0}
\definecolor{orange}{rgb}{0.8, 0.6, 0.2}
\definecolor{red}{rgb}{0.8, 0.2, 0.2}
\definecolor{darkred}{rgb}{0.6, 0.1, 0.05}
\definecolor{blueish}{rgb}{0.0, 0.3, .6}
\definecolor{light_gray}{rgb}{0.7, 0.7, .7}
\definecolor{pink}{rgb}{1, 0, 1}
\definecolor{greyblue}{rgb}{0.25, 0.25, 1}

\newcommand{\Tab}[1]{Tab.~\ref{table:#1}}
\newcommand{\Table}[1]{Table~\ref{table:#1}}

\usepackage{blindtext}

\renewcommand{\paragraph}[1]{\vspace{1em}\noindent\textbf{#1}.}
\begin{document}

\title{SCVRL: Shuffled Contrastive Video Representation Learning}

\author{Michael Dorkenwald$^1$\thanks{This work was done during an internship at AWS.}, $\;$ Fanyi Xiao$^2$, $\;$ Biagio Brattoli$^2$, $\;$ Joseph Tighe$^2$, $\;$ Davide Modolo$^2$ 
\\
$^1$Heidelberg University \quad $^2$AWS AI Labs \\
}
\maketitle


\begin{abstract}

\noindent We propose SCVRL, a novel contrastive-based framework for self-supervised learning for videos. Differently from previous contrast learning based methods that mostly focus on learning visual semantics (e.g., CVRL), SCVRL is capable of learning both semantic and motion patterns. For that, we reformulate the popular shuffling pretext task within a modern contrastive learning paradigm. 
We show that our transformer-based network has a natural capacity to learn motion in self-supervised settings and achieves strong performance, outperforming CVRL on four benchmarks. 


\end{abstract}

\section{Introduction}
\label{sec:intro}
\begin{figure}[t]
    \centering
    \includegraphics[width=0.5\textwidth]{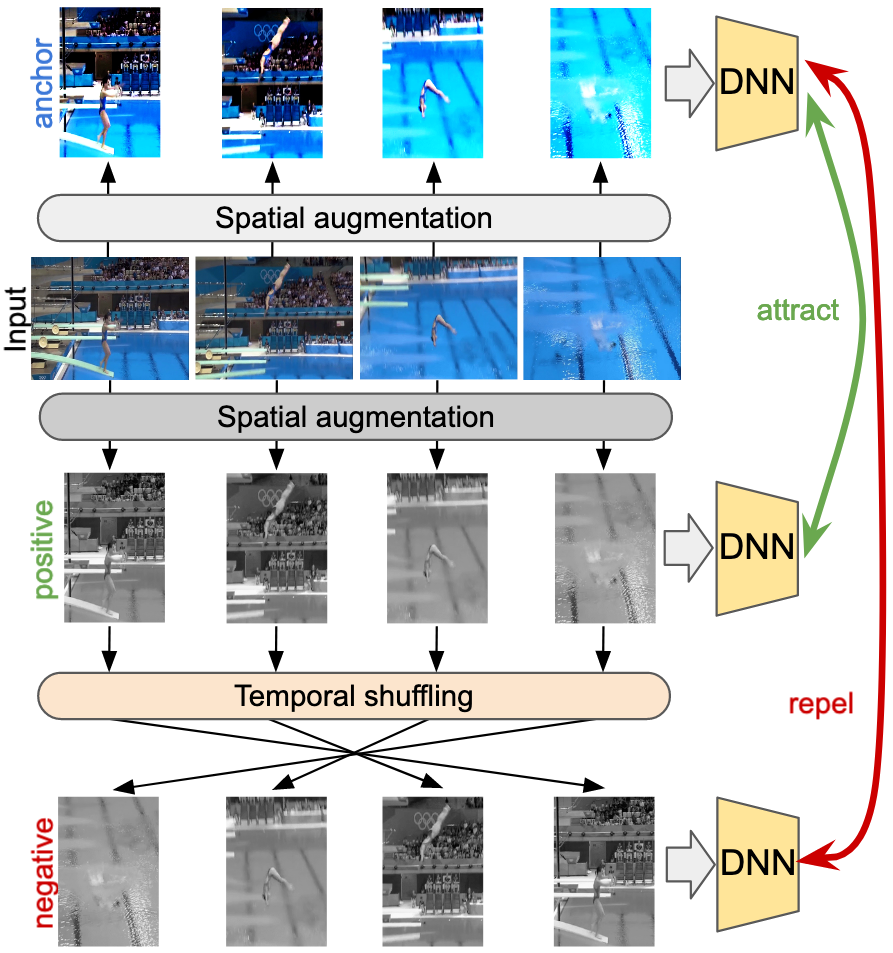}
    \caption{
    \textbf{Our temporal contrastive loss.} 
    We reformulate the popular pretext task of frame shuffling within a novel contrastive learning framework, where all the contrasted samples (anchor, positive and negative) are augmentations of the same clip (input). \vspace{-2mm}
    }
    \label{fig:firstpage}
\end{figure}
In recent years, self-supervised approaches have shown impressive results in the area of unsupervised representation learning~\cite{chen-simclr2020, grill-byol2020}. These methods are not bound to categorical descriptions given by labels (e.g., action classes) and thus are able to learn a representation that is not biased towards a particular target task. In this context, rich representations can be learned using pretext tasks~\cite{brattoli-cvpr2017, wei-cvpr2018, xu-vcop2019, vondrick-color2018,noroozi-jigsaw2016, gidaris-iclr2018, wang-cvpr2019} or, more recently, using contrastive learning objectives~\cite{hadsell-cvpr2006, wu-instdisc2018, he-moco2020, chen-simclr2020, chen-mocov2, chen-simsiam2020, misra-pirl2020, grill-byol2020, caron-swav2020,qian-cvrl2020} that led self-supervised performance to outperform the fully-supervised counterpart in the image domain~\cite{chen-simclr2020, grill-byol2020}. 

\begin{figure*}[!t]
    \centering
    \includegraphics[width=1\textwidth]{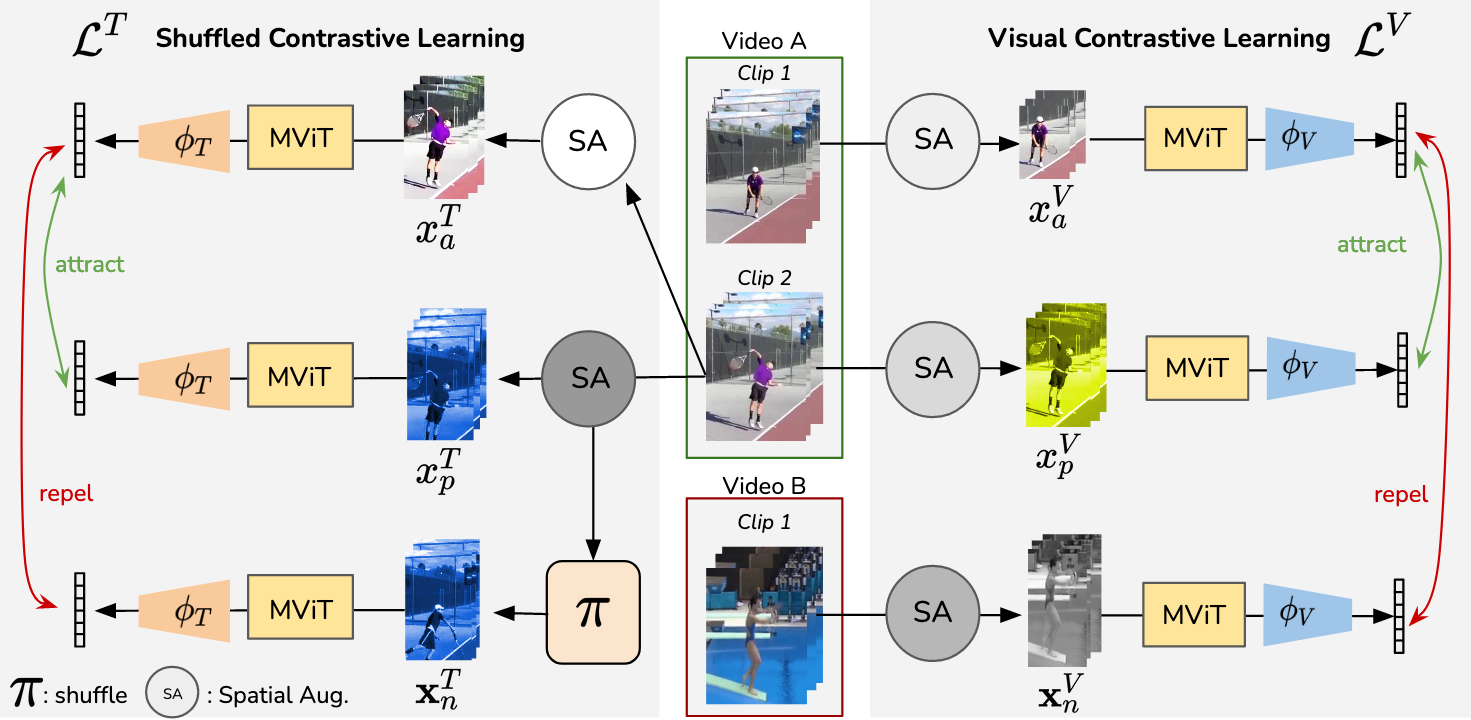}
    \caption{
    \textbf{Method overview.} Our framework consists of two contrastive objectives: \textit{(i)} our novel shuffled contrastive learning (left) is computed using the same clip (Video A, Clip 2) for all three samples, in fact anchor (top) and positive (middle) only differ by their spatial augmentation, then the positive is \textbf{shuffled} ($\pi$) over the time dimension to produce the negative sample (bottom); \textit{(ii)} visual contrastive learning (right), similarly to CVRL\cite{qian-cvrl2020} uses two clips from the same video (Video A) are used as anchor (Clip 1, top) and positive (Clip 2, middle) and a clip from a different video (Video B) constitutes the negative sample (bottom). Each sample is then run through the respective network composed of a shared backbone and two different heads, one for each loss. The visual contrastive loss extracts only semantic information and forces temporal invariance. The shuffled contrastive loss encourages the network to learn temporal information since it is the only characteristic that distinguishes positive from negative.
    }
    \label{fig:method}
\end{figure*}

While contrastive learning in images has advanced the state-of-the-art, it remains relatively under-explored in the video domain, where the few existing works applied contrastive learning to videos \cite{qian-cvrl2020, Hu_2021_ICCV_contrast_order, recasens-brave2021} in a way that primarily captures the semantics of the scene and disregards motion.
For example, CVRL~\cite{qian-cvrl2020} adapted the SimCLR image-based contrastive framework~\cite{chen-simclr2020} to videos by forcing two clips from the same video to have similar representations, while pushing apart clips from different videos (Fig.~\ref{fig:method}, right). 
This leads to strong visual features that are particularly effective on datasets such as UCF~\cite{soomro-ucf2012}, HMDB~\cite{kuehne-hmdb2011}, and Kinetics~\cite{kinetics}, where context and object appearance matter more than motion information. 
However, forcing the representations of two clips from a single video to be the same induces invariance to temporal information. For example, in a ``high jump'' video, this will force the video encoder to embed the ``running phase'' at the start of the video to use the same representation of the ``jumping phase'' at the end, even though these contain very different motion patterns. Instead, we propose a new  framework that can learn semantic-rich and motion-aware representations.

To accomplish this task, we borrow inspiration from the literature of self-supervised pretraining using pretext tasks for videos, like shuffle detection~\cite{xu-vcop2019, brattoli-cvpr2017, lee-iccv2017, buchler-eccv2018}, order verification~\cite{Hu_2021_ICCV_contrast_order} or frame order prediction~\cite{lee_frame_ordering_iccv, xu-vcop2019}. These methods learn by predicting a property of the video transformation (\eg the original order of the shuffled frames). While achieving competitive results, they learn representations that are covariant to their transformations~\cite{misra-pirl2020}, hindering their generalization potential for downstream tasks. Instead, we reformulate the previous shuffle detection pretext task in a contrastive learning formulation that yields stronger temporal-aware representations which learn motion beyond just frame shuffling and can therefore better generalize to new tasks. 


In detail, we propose SCVRL, a new contrastive video representation learning method that combines two contrastive objectives: a novel shuffled contrastive learning and a visual contrastive learning similar to \cite{qian-cvrl2020}. In order to encourage the learning of motion-sensitive features, our shuffled contrastive approach forces two augmentations of the same clip (anchor and positive, Fig.~\ref{fig:firstpage}) to have similar representations, while pushing apart negative clips generated by temporally shuffling the positive clip. Since all these samples come from the same original clip, during training the model cannot just look at their visual semantics to solve the contrastive objective, but is instead forced to reason about their temporal evolution, which is the key to the proposed learning.
%
Thanks to the rich combination of two contrastive objectives, SCVRL learns representations that are both motion and semantic-aware. SCVRL trains in an end-to-end fashion and its design consists of a single feature encoder with two dedicated MLP heads, one for each of the contrastive objectives.

As feature encoder, SCVRL adopts a Multiscale Vision Trasformer~\cite{MViT}, which is well suited for learning the rich SCVRL objective. 
We evaluate SCVRL on four popular benchmarks: Diving-48~\cite{diving-48}, UCF101~\cite{soomro-ucf2012} and HMDB~\cite{kuehne-hmdb2011} and Something-Something-v2~\cite{ssv2}. We show strong performance, consistently higher than the CVRL baseline on all datasets and metrics. Additionally, we also conduct an extensive ablation study to show the importance of our design choices and investigate the extent to which SCVRL learns motion and semantic representations. 

\section{Related Work}

\label{sec:related}

\paragraph{Self-supervised learning for images} 
Early works on self-supervised image representation learning used various pretext tasks such as image rotation prediction \cite{gidaris-iclr2018}, auto-encoder learning \cite{vincent-ICML-2008, song-ieee-2018, pathak-CVPR-2016}, or solving jigsaw puzzles \cite{noroozi-jigsaw2016} to learn semantic generalizable representations. 
That led to promising results, unfortunately still far away from those of fully-supervised models, mostly due to the difficulties of preventing the network from learning short-cuts. 
That changed with the (re-)emergence of approaches based on contrastive learning~\cite{hadsell-cvpr2006, wu-instdisc2018, he-moco2020, chen-simclr2020, chen-mocov2, chen-simsiam2020, misra-pirl2020, grill-byol2020, caron-swav2020}.
The underlying idea behind these approaches is to attract representations of different augmentations of the same image (positive pair) while repelling against those of different instances (negative pair)~\cite{bachman-neurips-2019, oord-arxiv2018}.
Thanks to this training paradigm, recent works were able to produce results on par with those of fully-supervised approaches~\cite{he-moco2020, chen-simclr2020, chen-mocov2, chen-simsiam2020, misra-pirl2020, grill-byol2020, caron-swav2020}.

\paragraph{Self-supervised learning for videos}
In the video domain, many works exploited the temporal structure of videos by designing specific pretext tasks such as pace prediction \cite{benaim-speednet2020, wang_pace_prediction}, order prediction \cite{misra-eccv2016, brattoli-cvpr2017, buchler-eccv2018, wei-cvpr2018, xu-vcop2019, lee_frame_ordering_iccv}, future frame prediction \cite{diba-dynamonet2019, Dorkenwald_2021_CVPR, Blattmann_2021_CVPR, Blattmann_2021_ICCV} or by tracking patches \cite{wang-iccv2015}, pixels \cite{wang-cvpr2019} or color \cite{vondrick-color2018} across neighboring frames. More recently, several approaches \cite{qian-cvrl2020, recasens-brave2021, Hu_2021_ICCV_contrast_order, fanyi_modist} adopted contrastive learning objectives from the image domain to learn stronger video representations. Among these, \emph{CVRL}~\cite{qian-cvrl2020} was the first to propose a video-specific solution for sampling pairs (positives and negatives) for contrastive video learning. In detail, CVRL proposed to sample positive pairs of clips from within the same video (their only constraint is that they cannot be too close to each other temporally) and negatives from other videos. This achieves impressive results on downstream tasks and it inspired this work (SCVRL). Specifically, it inspired us to design a novel sampling strategy that is even more suitable for contrastive video learning and that encourages the representation to learn \emph{both} semantic (as CVRL) and motion cues (using our novel shuffling contrastive learning).  
Finally, note how CVRL also inspired other very recent works~\cite{recasens-brave2021, Hu_2021_ICCV_contrast_order}. \cite{recasens-brave2021} tried to match the representation of a short clip to the one of a long clip, while \cite{Hu_2021_ICCV_contrast_order} tried to incorporate temporal context using an order verification pretext task. We argue that none of these approaches explicitly enforce the representation to learn motion information like our SCVRL.

\paragraph{Vision Transformers} In recent years, transformer models~\cite{vaswani_attention_NIPS17, BERT_Devlin} have achieved unprecedented performance on many NLP tasks. Dosovitskiy et al. \cite{dosovitskiy_vision_transformer_ICLR21} adapted them to the image-domain with a convolution-free architecture (ViT) achieving competitive results on image classification and sparking a new research trend in the field. Since then, many works have been proposed, including some in the video domain \cite{timesformer, MViT, arnab_vivit_2021, zhao_tighe_TubeR}. 
Among these, MViT~\cite{MViT} is of particular interest to this paper, as we use it as the feature encoder for SCVRL. MViT proposes a multiscale feature hierarchy for transformers to effective model dense visual inputs without the need for external data. It produces state-of-the-art results on various video classification benchmarks and we believe its architecture design is well suited to learn the contrastive objective of SCVRL.
\section{Method}
\label{sec:method}
%


SCVRL is a novel contrastive learning framework that learns representations which are \emph{both} rich in semantics and sensitive to motion cues. To achieve this, it leverages two objectives: a classic \emph{visual} contrastive objective to learn the semantics of the video (e.g., CVRL~\cite{qian-cvrl2020}) and 
a novel objective which compares a video clip and the same clip \emph{temporally shuffled} (Fig.~\ref{fig:firstpage}), such that the learnt representation is aware of the temporal order of an action and can distinguish its different phases.

\subsection{Preliminaries on Contrastive Learning}
The objective of contrastive learning is to produce an embedding space by attracting positive pairs, $x_a$ (anchor) and $x_p$ (positive), while pushing away a set of $N$ negatives $\mathbf{x_n}= \{x_{n_1}, \dots, x_{n_N}\}$. This is achieved by training an encoder $f$ which embeds a given video clip $x$ to a $\mathbb{L}_2$ normalized feature vector using the InfoNCE~\cite{oord-arxiv2018,he-moco2020} loss $\mathcal{L} = \text{IN}(f; x_a,x_p,\mathbf{x}_n)$ which is formulated as follows:
\begin{align}
\mathcal{L} = -\log \frac{e^{f(x_a)^T{\cdot}f(x_p) / \tau}}{e^{f(x_a)^T{\cdot}f(x_p) / \tau} + \sum_{i=1}^{N}e^{f(x_a)^T{\cdot}f(x_{n_i})  / \tau}}, 
\label{eq:lt}
\end{align}
with $\tau>0$ as a temperature parameter. 
It is clear from this formulation that what the model learns largely depends on how the positive and negative pairs are sampled. 
In the next section we delve into how we construct training samples for our shuffled contrastive learning objective and in the following one we then present our SCVRL framework.

\subsection{Shuffled Contrastive Learning}
\label{sec:temporal_contrastive}


\begin{figure}[!t]
    \centering
    \includegraphics[width=0.48\textwidth]{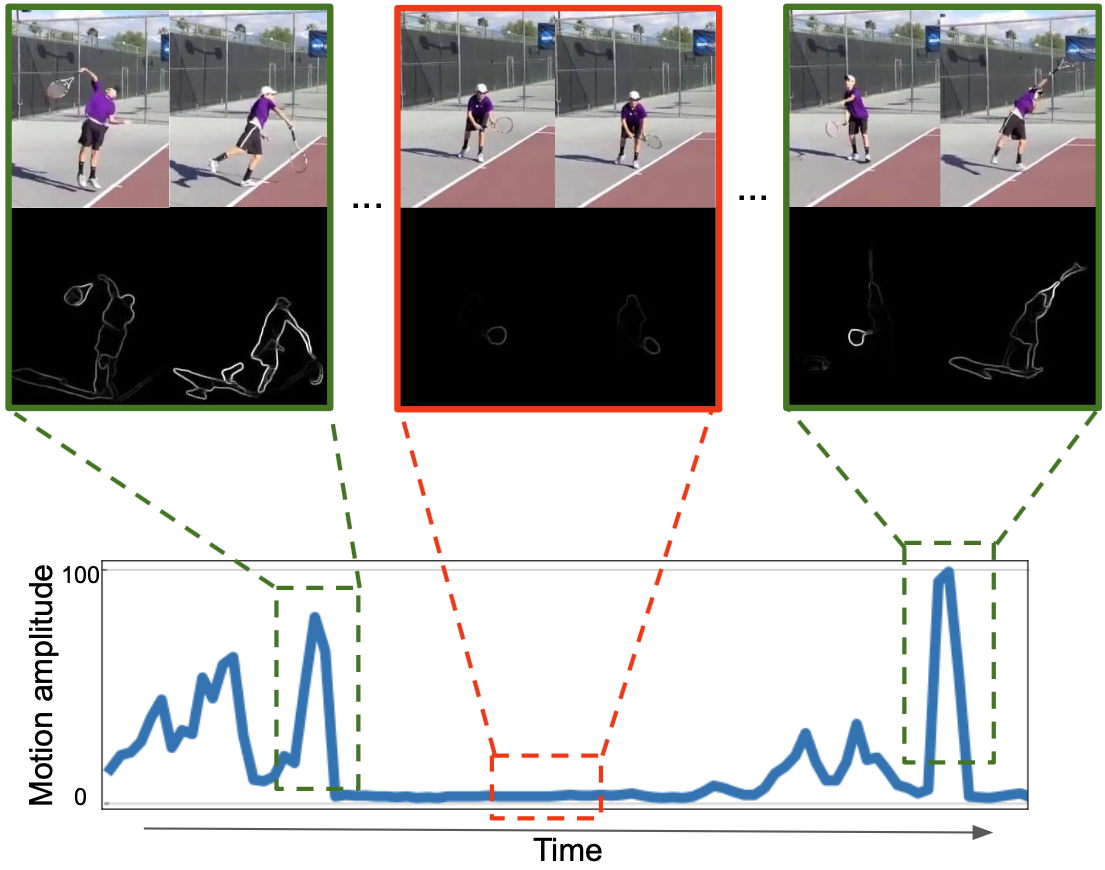}
    \caption{
    \textbf{Targeted Sampling.} During self-supervised training, we select clips with high motion. For each video, we compute the frame difference (middle), select the top 4K pixels and calculate the media over 1 second (bottom plot). In this example, the first and last clip (green boundaries) are selected given that the tennis player is moving (hitting the ball). The middle clip (red boundaries) is excluded from self-supervised training since the subject is only preparing the ball and not moving, therefore shuffling these frames would not produce different enough samples to learn motion information.}
    \label{fig:targeted_sampling}
\vspace{-0.5cm}
\end{figure}
The key to ensure that $\mathcal{L}$ learns a motion-aware representation space lies in how positive, negative, and anchor clips are sampled during training. 
Differently from previous approaches (e.g. CVRL~\cite{qian-cvrl2020}), that sample positive pairs from one video and negatives from other videos, 
we use a single clip to produce anchor, positive and negatives and only change the temporal order of its frames as a mean to learn temporal sensitive features.

Specifically, we extract a single clip from a video and apply two different spatial augmentations to obtain our positive pair $x_a^T$ and $x^{T}_p$ (Fig.~\ref{fig:method}). Our negatives are then generated by applying a series of temporal permutations $\pi_i \in \Pi$ to the positive clip: $x^T_{n_i}=\pi_i(x^{T}_p)$ (i.e., \emph{shuffling}). In the context of Eq.~\ref{eq:lt}, we define our temporal shuffled contrastive objective as 
$\mathcal{L}^T = \text{IN}(f; x^T_a, x^T_p, \mathbf{x}_n)$.
This objective forces the encoder to push apart the representations of the anchor clip $x_{a}^T$ (which is in normal order) and the shuffled clip $\mathbf{x}^T_n$.
This makes the learned representation sensitive to frame ordering (\ie, motion-aware), as these pairs $x_{a}^T$ and $\mathbf{x}^T_n$ share the same semantics and the model can only repel them using motion information.
Furthermore, in order to avoid trivial solutions, an important design choice is to make sure $x^T_{n_i}$ and $x_a^T$ do not share the same spatial extent. 
To achieve this, we generate $x^T_{n_i}$ by permuting $x^{T}_p$ rather than $x^{T}_a$. 
This critical design avoids the shortcut where the encoder solves the task with a simple pixel-level comparison between frames.

At the same time, this design also carries a potential risk: if $x_{p}^T$ is a clip containing almost no motion (e.g., static scenes), then its shuffled representation $x^T_{n_i}$ would look identical to it and the encoder would not be able to push them apart and learn correctly. To overcome this problem and take full advantage of our shuffling contrastive learning method, we propose a new targeted sampling approach that aims at selecting the right clips for shuffling. 

\paragraph{Probabilistic targeted sampling} 
Given a video, we want to sample clips that have strong motion for shuffling, as visualized in Fig.~\ref{fig:targeted_sampling}. To measure the motion of each clip, we use optical flow edges~\cite{fanyi_modist}, which have been shown to be more robust to global camera motion compared to raw optical flow. We estimate flow edges by applying a Sobel filter~\cite{sobel-sobel2014} onto the flow magnitude map and take the median over the highest 4k flow edge pixels values in each frame. Then, we aggregate them in time by taking the median over all the frames in a 1-second temporal window ($m_i$).
Rather than deterministically sampling the clip with the highest motion (which would disregard a lot of useful clips), 
we propose to sample from a multinomial distribution based on the probability $p_i$ computed from window $i$ by inserting its motion amplitude $m_i$ into a softmax function:
\begin{align}
    p_i = \frac{e^{m_i / \beta}}{\sum_{j}^{C} e^{m_j / \beta}},
\label{eq:softmax}
\end{align}
where $\beta$ is the temperature which regularizes the entropy of the distribution and $C$ is the number of temporal windows inside the video. 

\newcommand{\blockMViT}[3]{\multirow{2}{*}{\(\left[\begin{array}{c}\text{{#2}}\\[-.1em] \text{#1}\end{array}\right]\)$\times$#3}
}
\begin{table}[t]
	\centering
	\footnotesize
	\resizebox{0.78\columnwidth}{!}{
		\tablestyle{1pt}{1.08}
		\begin{tabular}{c|c|c}
			stages & operators &  output sizes \\
			\shline
			\multirow{1}{*}{data layer} & stride $4\times 1 \times 1$ & 3\x16\x224$^2$ \\
			\hline
			\multirow{2}{*}{cube$_1$}  & \multicolumn{1}{c|}{$\mathbf{2}\times 7 \times 7, 96$} & \multirow{2}{*}{$96 \times 8 \times 56^2$} \\
			& stride $2\times 4\times 4$ &  \\
			\hline
			\multirow{2}{*}{scale$_2$} & \blockMViT{{MLP(384)}}{{MHPA(96)}}{1} & \multirow{2}{*}{$96 \times 8 \times 56^2$} \\
			&  & \\
			\hline
			\multirow{2}{*}{scale$_3$} & \blockMViT{{MLP(768)}}{{MHPA(192)}}{2} &  \multirow{2}{*}{$192 \times 8 \times 28^2$} \\
			&  & \\
			\hline
			\multirow{2}{*}{scale$_4$} & \blockMViT{{MLP(1536)}}{{MHPA(384)}}{11} & \multirow{2}{*}{$384 \times 8 \times 14^2$} \\
			&  & \\
			\hline
			\multirow{2}{*}{scale$_5$} & \blockMViT{{MLP(3072)}}{{MHPA(768)}}{2} & \multirow{2}{*}{$768 \times 8 \times 7^2$}\\
			&  & \\
			\hline
			\multicolumn{3}{c}{Output: CLS token} \\
		\end{tabular}
		}
	\caption{\small  \textbf{\ours{} backbone architecture}. The network is the same as MViT-B~\cite{MViT} except for the first layer (cube$_1$) where the temporal kernel size is reduced to avoid shortcuts (see Sec. \ref{sec:impl_details}). The dimensions of the output size is denoted as $C \times T \times H \cdot W$. }
	\label{table:arch}
\vspace{-0.5cm}
\end{table}

\subsection{SCVRL Framework}
\label{sec:framework}
Our self-supervised SCVRL framework learns a video representation that is both semantic-rich and motion-aware. To achieve this, it combines two contrastive objectives: our {\bf T}emporal shuffled contrastive learning objective $\mathcal{L}^T$ and a {\bf V}isual contrastive learning objective $\mathcal{L}^V$:
\begin{align}
    \mathcal{L}^{F} = \mathcal{L}^T + \lambda \mathcal{L}^V,
\label{eq:SCVRL}
\end{align}
where $\lambda$ is a weighting parameter and $\mathcal{L}^V$ is the contrastive objective used in CVRL~\cite{qian-cvrl2020} that helps SCVRL learn semantic features. 
Its objective is illustrated in Fig.~\ref{fig:method} (right) and is defined as: $\mathcal{L}^V = \text{IN}(f; x^V_a, x^V_p, \mathbf{x}^V_n)$, where $x^V_a=x^T_a$, $x^V_p$ is a different clip from the same video of $x^V_a$ and the negative clips $\mathbf{x}^V_n$ are sampled from entirely different videos. 
To the best of our knowledge, \ours{} is the first work that explicitly models both motion and visual cues within the same self-supervised contrastive learning framework.
%
Finally, we note that one cannot naively combine these two competing objectives as $\mathcal{L}^V$ wants to pull together the representation of all the clips within a video, while $\mathcal{L}^T$ wants to push them away when shuffled.  To circumvent this, we design SCVRL with a shared backbone encoder $f$, but two independent MLP heads $\phi_V$ and $\phi_T$, one for each of the objectives (\Table{ablation-heads}, Fig.~\ref{fig:method})~\cite{milbich_diva_ECCV, sharing_PAMI}. 
For the backbone video encoder, we adopt Multiscale Vision Transformer~\cite{MViT} (MViT), which has a large temporal receptive field that makes it particularly suitable for learning motion cues using our shuffled contrastive learning objective.

%

\begin{table}[t]
  \centering
  \small
  \begin{tabular}{l | c | c | c}
  \hline
    \multicolumn{1}{c|}{Method}  & \hspace{0.3mm} Shuffle \hspace{0.3mm} & \hspace{1mm} UCF & \hspace{1mm} Diving-48
    \\
    \hline
    \hline
    Rand Init &  \xmark & \hspace{1mm} 13.4 {\color{White} (-- 6.7)} & 9.5 {\color{White}(--6.7)}
    \\
    Rand Init & \cmark & \hspace{1mm} 12.8 {\color{Red} (--0.6)}  & 8.7 {\color{Red}(--0.8)}
    \\
    \hline
    CVRL & \xmark & \hspace{1mm} 67.4 {\color{White} (--6.7)} & 12.0  \textbf{\color{White} (--6.7)}
    \\
    CVRL  &  \cmark & \hspace{1mm} 62.9 {\color{Red} (--3.5)} &  9.5  {\color{Red} (--2.5)}
    \\
    \hline
    SCVRL & \xmark & \hspace{1mm} 68.0  {\color{White}(--6.7)} & 11.9 {\color{White}(--3.7)}
    \\
    SCVRL  & \cmark & \hspace{1mm} 56.2 \textbf{\color{Red} (--11.8)} & 8.1  \textbf{\color{Red} (--3.8)}
        \\
    \hline
  \end{tabular}
  \caption{\small \textbf{Dependency on motion information.} 
  We evaluate the performance drop when shuffling the input during inference as introduced by \cite{MViT, sevilla_wacv_time_can_tell}. A large drop indicates that the network relies on temporal information to solve the task. A small drop, on the other hand, the model is basing the decision only on per-frame semantic. The models are trained using the linear evaluation protocol for the corresponding dataset and we report top-1 accuracy. The performance drop is indicated in red brackets.
  }
  \label{table:shuffle_ablation}
\end{table}

\section{Experiments}

In this section, we first describe the implementation details and the benchmark datasets we use in our experiments. Then, we investigate the motion information extracted by our \ours{} compared to the baseline. Finally, we evaluate our model on the downstream task of action recognition and perform several ablation studies to better understand the impact of our design choices.

\subsection{Implementation Details} \label{sec:impl_details}
\noindent \textbf{Training protocol.}
We pretrain \ours{} on the Kinetics-400 (K400) dataset~\cite{kay-kinetics2017}, which contains around 240k 10-seconds videos, \textit{without} the use of the provided action annotations.
Our final model is pretrained on K400 for 500 epochs. Since the training of transformers is computationally very expensive, we validate the strong performance of our model by comparing to the CVRL baseline on a training scheme using 100 epochs.
For efficiency, we only train for 50 epochs for our ablation study on a subset of K400 containing 60k videos that we call \textit{Kinetics400-mini}. 
We train \ours{} with a learning rate of $1 \times 10^{-4}$ with linear warm-up and cosine annealing using AdamW optimizer~\cite{Loshchilov_adamw_ICLR2019} with a batch size of 4 per GPU and weight decay of $0.05$. We set the warm-up and the end learning rate to $1\times 10^{-6}$. 
We set $\lambda$ in Eq. \ref{eq:SCVRL} to 1. The temperature $\beta$ for our targeted sampling is set to 5 as ablated in Fig. \ref{table:ablation-sampling}.
The spatial augmentations are generated with random spatial cropping, temporal jittering, $p$ = 0.2 probability grayscale conversion, $p$ = 0.5 horizontal flip, $p$ = 0.5 Gaussian blur, and $p$ = 0.8 color perturbation on brightness, contrast and saturation, all with 0.4 jittering ratio. The same augmentation is applied to all frames within a clip.
To effectively train SCVRL with $\mathcal{L}^V$ we follow~\cite{wu-instdisc2018, he-moco2020} and construct a \emph{memory bank} of $N^V$ negative samples. To train with $\mathcal{L}^T$, instead, we generate negatives on the fly by randomly permuting the positive clip $N^T$ times. For the visual contrastive objective, we use a memory bank size $N^V$ of 65536 negative samples, while for the temporal counterpart we set $N^T$ to $12$. For both objective we set the temperature $\tau$ to 0.1. As suggested in \cite{he-moco2020} we maintain a momentum version of our model and process anchor clips with our online model while positive and negat clips 
are processed with the momentum version.

\begin{table}[t]
\resizebox{\linewidth}{!}{
  \centering
  \begin{tabular}{ l | c | c }
  \hline
    Category & Motion & $\Delta$ Acc. $\uparrow$
    \\
    \hline
    \hline
    Turning the camera downwards while filming sth & 0.69 & 22.5 
    \\
    Turning the camera left while filming sth & 0.95 & 17.8
    \\
    Digging sth out of sth & 0.64 & 12.1 
    \\
    Turning the camera upwards while filming sth & 0.89& 11.6 
    \\
    Uncovering sth & 0.65& 10.8 
    \\
    \hline
    Showing a photo of sth to the camera  & 0.25 & -1.9 
    \\
    Showing sth on top of sth & 0.08& -2.6 
    \\
    Scooping sth up with sth & 0.43 & -2.9 
    \\
    Pulling two ends of sth so that it gets stretched  & 0.43 & -3.3
    \\
    Throwing sth against sth & 0.42& -5.0 
    \\
    \hline
  \end{tabular}
  }
  \caption{
  \small  \textbf{Performance gain related to motion.} 
  For each SSv2 action class, we compare the absolute difference in accuracy between SCVRL and CVRL ($\Delta$ Acc.) using linear evaluation and the normalized motion.
  For this table, we show five of the highest and lowest performing classes, sorted by $\Delta$ Acc. 
  The result shows that \ours{} improves on actions with high motion by a large margin. At the same time, CVRL outperforms on classes with low motion by only a marginal difference.
  }
  \label{table:ssv2_comparison}
\vspace{-0.4cm}
\end{table}

\begin{figure*}[!t]
    \centering
    \includegraphics[width=1\textwidth]{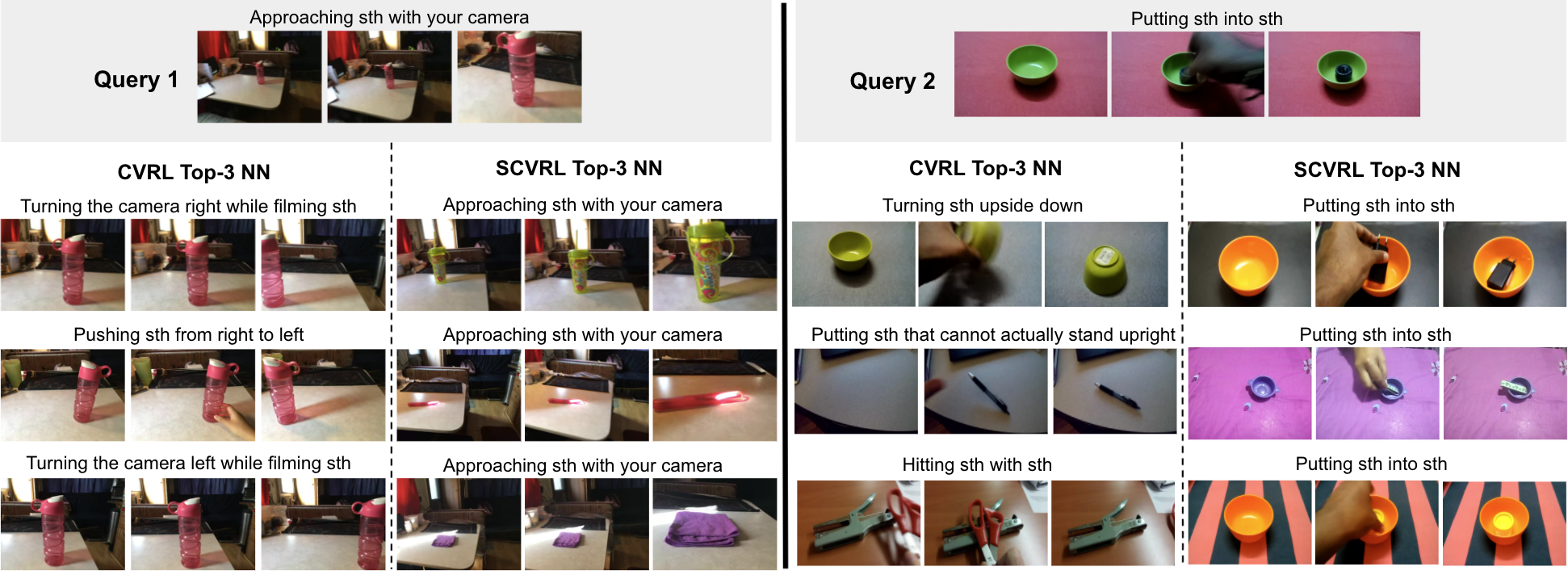}
    \caption{
    \textbf{Video retrieval comparison.} For each query, we show the top three nearest-neighbor based on CVRL and SCVRL representations. For each video, we show the first, middle and last frame. The ground-truth class is provided on top. 
    The figure illustrates that CVRL is biased towards appearance while SCVRL is aware of motion patterns. For example, in Query 1 CVRL retrieves videos of the same scene even if the motion (i.e. action class) is different, while SCVRL retrieves nearest neighbors with different appearance but similar motion. 
    }
    \label{fig:nn}
\end{figure*}

\paragraph{Architecture} We use the \emph{MViT-Base} (MViT-B) version of MViT and operate on clips of 16 frames which are extracted with a stride of 4. Our architecture is shown Tab.~\ref{table:arch}. 
As in previous works~\cite{MViT, arnab_vivit_2021} we use a cube projection layer to map the input video to tokens. This layer is designed as a 3D convolution with a temporal kernel size of 3 and stride of 2 with padding of 1. 
However, when the input sequence is shuffled, the projection layer has direct access to the seams between the shuffled frames. This would allow the network to learn shortcuts by directly detecting if a sequence is shuffled. We ablated this behavior in Tab.~\ref{table:ablation-kernel}. 
To overcome this issue, we propose to shuffle groups of two frames and use a temporal kernel size to 2 with a stride of 2, thus avoiding convolutional kernels overlapping across shuffled tokens. 
Finally, each MLP head in SCVRL is a two-layer network that consists of a linear layer that transforms the CLS token to a 2048 dimensional feature vector, followed by a ReLU activation function, and a second linear layer that maps to an embedding of $128$-D.


\paragraph{Evaluation protocol and Baselines} Following~\cite{qian-cvrl2020, recasens-brave2021}, we employ two evaluation protocols to quantify self-supervised representations: (i) \textit{Linear}: Training a linear classification layer on top of the frozen pretrained backbone (ii) \textit{Full}: finetuning the whole network in an end-to-end fashion on the target dataset. For all evaluations we report the top1 accuracy. 
We follow previous works and use a standard of 10 temporal and 3 spatial crops during testing. 
We compare against two baselines: (i) {\it CVRL}~\cite{qian-cvrl2020}, a state-of-the-art self-supervised contrastive learning method trained using the visual contrastive loss $\mathcal{L}^{V}$ of Eq.~\ref{eq:SCVRL}; and (ii) {\it Supervised}, which is a fully supervised model trained for actions on K400. All baselines use the same MViT-B backbone as \ours{}.
\subsection{Datasets}\label{sec:dataset}

\noindent{\bf
Diving-48~\cite{diving-48}} is a fine-grained action dataset capturing 48 unique diving classes. It has around 18k trimmed video clips, each containing a diving sequence that consists of a mix of takeoff, the motion during the dive and water entering. It is a challenging dataset due to the semantic similarity across all diving classes (e.g., similar foreground and background, similar diving outfits, etc.). As the original Diving-48 dataset had some annotation issues, we follow ~\cite{timesformer} and use their cleaned annotations.

\noindent{\bf
Something-Something-v2~\cite{ssv2}} (SSv2), similarly to Diving-48, is a benchmark that was specifically developed to evaluate a model's capability to learn temporal dynamics. SSv2 consists of video clips showing complex human-object interactions, such as “Moving something up” and “Pushing something from left to right”. It contains a total of 174 unique actions, 168k training videos, 24k validation videos and 24k test videos.

\noindent{\bf UCF101~\cite{soomro-ucf2012} and \textbf{HMDB}~\cite{kuehne-hmdb2011}} are both standard benchmark datasets for the classification of sport activities and daily human actions. UCF contains 13320 YouTube videos labelled with 101 classes, while HMDB contains 6767 video clips labelled with 51 actions. Differently from the previous two datasets, these can reliably be solved using mostly semantic cues, as their action classes are quite distinct. For all experiments on UCF101 and HMDB51, we report results using split1 for train/test split.

\subsection{Analysis of Motion Information}
\label{subsec:analysis_motion}
The goal of our shuffled contrastive learning objective is to learn stronger temporal features.
We now quantify the amount of temporal information learned by \ours{} in comparison to CVRL.

\paragraph{Dependency on motion information} First, we investigate to what extend \ours{} has learnt to distinguish temporally coherent sequences compared to shuffled ones. For this, we follow previous works~\cite{MViT, sevilla_wacv_time_can_tell} and compute the performance change between running inference on a test clips \vs its shuffled version. Results using the Linear evaluation protocols are presented in \Tab{shuffle_ablation}. These show that SCVRL is very sensitive to the temporal order of the frames and its performance drops by $\sim12$ points Top-1 accuracy on UCF. This validates the effectiveness of our shuffled contrastive learning strategy. On the contrary, CVRL's performance on UCF is barely affected (-3.5), confirming that CVRL's representations are temporal invariant when trained on datasets that mostly rely on semantic cues (i.e., UCF). 

\paragraph{Correlation between motion quantity and performance}
We now compare per-class performance change between \ours{} and the baseline CVLR in Tab.~\ref{table:ssv2_comparison} and correlate it to the amount of motion each action contains. For this, we calculate the median motion magnitude in the pixel space for all videos of each class. 
We show five high (top) and five low performing (bottom) classes sorted by their performance gains.
First and foremost, the improvement brought by SCVRL is substantial (top: 10-20 Top-1 accuracy points) compared to its loss (bottom: 2-5 points). Second, \ours{} improves particularly on actions with large motion, like ``Turning the camera left'', showing that it is important to model motion cues in video representations and that \ours{} is capable to do that well.

\begin{table}[t]
  \centering
  \small
  \begin{tabular}{@{\hskip1pt}c@{\hskip4pt} l | c | @{\hskip0pt}c@{\hskip0pt} | c c}
    \hline
    &  Method & Pretrain data & \hspace{0.3mm} Sup. \hspace{0.3mm} &  Full &  Linear
    \\
    \hline
    \hline
    \multirow{4}{*}{\rotatebox[origin=C]{90}{\footnotesize{100 Epochs}}} 
    &Rand Init & -- &  \xmark &  18.4 &  6.9
     \\
     &CVRL & K400 & \xmark &  52.6 &  \textbf{12.1}
    \\
     &\ours & K400 & \xmark &  \textbf{53.8} &  11.9
    \\ 
    & \demph{MViT} & \demph{K400} & \demph{\cmark} &  \demph{68.8} &  \demph{22.3}
    \\
    \hline
    &\ours & K400 & \xmark &  66.4 &  18.1
    \\
    \hline
  \end{tabular}
  \caption{\small  \textbf{Action classification on Diving-48.} We pretrain on K400 and then transfer the representation to Diving48 for finetuning under both Linear and full protocols. Results are reported as top-1 accuracy. \ours{} outperforms the CVRL baseline by a significant margin. For context, we also compare to supervised video models pre-trained on Imagenet-1K and show that our model outperforms them.
  }
  \label{table:diving}
 \vspace{-0.1cm}
\end{table}

\begin{table}[t]
  \centering
  \resizebox{\linewidth}{!}{
  \small
  \begin{tabular}{@{\hskip1pt}c@{\hskip4pt} l | c| c | @{\hskip0pt}c@{\hskip0pt}  | cc | cc}
  \hline
    & \multirow{2}{*}{Method} & \multirow{2}{*}{Pretrain} & \multirow{2}{*}{Arch.} & \hspace{0.1mm} \multirow{2}{*}{Sup.} \hspace{0.1mm} & 
    \multicolumn{2}{c}{{UCF}} & \multicolumn{2}{c}{{HMDB}}\\
      & & & & & Full & Linear
        & Full & Linear \\
    \hline
    \hline
    \multirow{4}{*}{\rotatebox[origin=C]{90}{\footnotesize{100 Epochs}}}
    &Rand Init & K400 & MViT-B &  \xmark &  68.0 & 18.7 & 32.4 & 13.6
    \\
     &CVRL & K400 &MViT-B &  \xmark &  83.0 & 67.4 & 54.6 & \textbf{42.4}
    \\
     &SCVRL &K400 & MViT-B & \xmark &  \textbf{85.7} & \textbf{68.0}&  \textbf{55.4} & 40.8
    \\ 
     & \demph{Supervised} &\demph{K400} & \demph{MViT-B} & \demph{\cmark} &  \demph{93.7} & \demph{93.4}& \demph{68.8} & \demph{67.6}
     \\
    \hline
     &SCVRL & K400 &MViT-B & \xmark & 89.0 &  74.4 & 62.6 & 50.1
    \\
    & \demph{CVRL~\cite{qian-cvrl2020}}  & \demph{K400} & \demph{R3D-50}  & \demph{\xmark} & \demph{92.9}  & \demph{89.8} &  \demph{67.9} & \demph{58.3}
    \\
    \hline
  \end{tabular}
  }
  \caption{\small  \textbf{Action classification on UCF and HMDB.} We pretrain on K400 and then transfer the representation to UCF101 and HMDB for Linear and Full finetuning. Results are reported as top-1 accuracy. 
  }
  \label{table:ucf_hmdb}
\vspace{-0.5cm}
\end{table}

\paragraph{Video retrieval comparison}
In Fig.~\ref{fig:nn}, we compare the nearest neighbors obtained using the representation space learned by \ours{} to the one from CVRL on the SSv2 dataset. For each query we show the top-3 nearest neighbors. For both queries we observe that \ours{} better captures the temporal information in the query and the top-3 nearest neighbor are very similar in motion while having a large variation in semantics. In Query 1, the nearest neighbor from CVRL have all the same appearance, yet, completely different progressions. Also in Query 2 CVRL top-1 nearest neighbor displays a similar object as the query (green cup) but different motion, while SCVRL correctly retrieves a clip with the same motion even if the cup has a different color. 
This illustrates that the learned features from \ours{} better captures the missing temporal information in CVRL.

\subsection{Downstream Evaluation}

Next, we evaluate \ours{} on four datasets introduced in Sec.~\ref{sec:dataset} for action recognition, using the two evaluation protocols of Sec.~\ref{sec:impl_details}.

\begin{table}[t]
  \centering
  \small
  \begin{tabular}{ @{\hskip1pt}c@{\hskip4pt} l | c | c | @{\hskip1pt}c@{\hskip1pt} | c | c}
  \hline
    & Method & Arch. &Pretrain &  Sup. &  Full &  Linear 
    \\
    \hline
    \hline
    \multirow{4}{*}{\rotatebox[origin=C]{90}{\footnotesize{100 Epochs}}} 
    &Rand Init & MViT-B & -- &  \xmark &  38.7 &  3.2
    \\
     &CVRL & MViT-B & K400 & \xmark &  45.3 &  11.4
    \\
     &\ours & MViT-B& K400 & \xmark &  \textbf{46.8} &  \textbf{13.8}
     \\ 
     & \demph{MViT} & MViT-B & \demph{K400} & \demph{\cmark} &  \demph{53.7} &  \demph{19.4}
    \\
    \hline
      & \ours & MViT-B & K400 & \xmark &  53.5 & 19.4
    \\
    & \demph{MViT \cite{MViT}} & \demph{MViT-B}& \demph{K400} & \demph{\cmark} &  \demph{64.7} &  \demph{--}
         \\
    \hline
  \end{tabular}
  \caption{\small  \textbf{Action classification on SSv2.} We pretrain on K400 and then transfer the representation to SSv2 for finetuning under both Linear and Full protocols. Top-1 accuracy is reported in the table. 
  }
  \label{table:ssv2}
\vspace{-0.2cm}
\end{table}

\newcolumntype{L}{>{\raggedright\arraybackslash}X}
\newcolumntype{C}{>{\centering\arraybackslash}X}

\begin{table}[t]
\resizebox{\linewidth}{!}{
	\centering
	\small
	\begin{tabular}{l | c c c c c c}
	\multicolumn{1}{l|}{Method} 		& 1\% 		& 5\% 				& 10\% 			& 25\% 			& 50\% 			& 100\%
		\\
	\shline
    \demph{Rand Init}    &  \demph{1.5}   &  \demph{2.9}         & \demph{7.8}  & \demph{18.7}  & \demph{28.5}  & \demph{40.0}  
    \\
		\hline
		CVRL  	& 4.4	& 15.4 	& 23.3 	& 31.0 	& 40.1 	& 45.3 
		\\
		SCVRL 		& 6.1 		& 19.4 	& 26.3 	& 34.4 	& 43.0 	& 46.8 	
		\\
		rel. $\Delta \%$ & +27.9& +20.6	& +11.4 & +9.9	& +6.7	& +3.2	
		
	\end{tabular}
	}
	\caption{\small  \textbf{Low-shot learning on SSv2.} Rows indicate different pretraining method on K400, while columns vary the \% of SSv2 training data used for finetuning. All results are top-1 accuracy. Our method (\ours{}) consistently provides an higher gain than CVRL and, in particular, achieves higher gain when the amount of annotations are lower.}
	\label{table:lowshot}
\vspace{-0.5cm}
\end{table}
\paragraph{ Action Classification on Diving-48}  
\Tab{diving} shows that our \ours{} achieves better performance than baseline for full finetuning on Diving-48 while being on par for the linear evaluation protocol.


\begin{table*}[t]\centering 
  \captionsetup[subfloat]{captionskip=2pt}
  \subfloat[\textbf{Targeted sampling}
  \label{table:ablation-sampling}]{
    \tablestyle{2pt}{1.05}
    \begin{tabular}{c | c c }
      \multicolumn{1}{c|}{Temp. $\beta$}  & UCF & SSv2  
      \\
      \shline
      $\infty$ & 75.0 & 43.8
      
      \\
      10  & 75.1 & 43.9
      \\
      \textbf{5}  & \textbf{75.5} & \textbf{44.3} 
      \\
     3  & 74.1 & 43.1 
  \end{tabular}}\hspace{2mm}
  \subfloat[\textbf{Contrastive heads}
  \label{table:ablation-heads}]{
    \tablestyle{2pt}{1.05}
    \begin{tabular}{c | c c }
      \multicolumn{1}{c|}{Method} & UCF & SSv2
      \\
      \shline
      CVRL &  70.3 & 41.9
      \\
      SCVRL shared & 69.6 & 40.7
      \\
      \textbf{SCVRL separate} & \textbf{75.5} & \textbf{44.3}
      \\
  \end{tabular}}\hspace{2mm}
  \subfloat[\textbf{Ablate $\mathcal{L}_{T}$} \label{table:ablation-components}]{
    \tablestyle{2pt}{1.05}
    \begin{tabular}{l | c c c}
      Method & UCF & SSv2 
      \\
      \shline
      CVRL & 70.3 & 41.9
      \\
      $+$ cls shuffle & 70.3 & 42.6
      \\
      \textbf{\ours{}} & \textbf{75.5} & \textbf{44.3}
  \end{tabular}}\hspace{2mm}
  \subfloat[\textbf{CLS vs AVG pooling} \label{table:ablation-representation}]{
    \tablestyle{2pt}{1.05}
    \begin{tabular}{c | c c }
      Method & UCF & SSv2 
      \\
      \shline
      CVRL & 70.3 & 41.9
      \\
      \textbf{\ours{} CLS} & \textbf{75.5} & \textbf{44.3} 
      \\
      \ours{} AVG & 74.2 & 43.9
  \end{tabular}}\hspace{2mm}
    \subfloat[\textbf{Proj. layer kernel size} \label{table:ablation-kernel}]{
    \tablestyle{2pt}{1.05}
    \begin{tabular}{c | c| c c }
      Method & $k_t$ & UCF & SSv2 
      \\
      \shline
      CVRL & 2 & 70.3 & 41.9
    \\
      \textbf{\ours{}} & 2 & \textbf{75.5} & \textbf{44.3} 
      \\ \hline
      CVRL & 3 & 72.6 & 41.7
      \\
      {\ours{}} & 3 &{72.7} & {41.9} 
  \end{tabular}}
  \caption{\small  \textbf{Ablating \ours{}}. We present top-1 classification accuracy using the Linear evaluation protocol on UCF and SSv2. For efficiency, we train the corresponding model on K400-mini since we are looking at relative improvements. \vspace{2mm}}
  \label{table:ablation-all}
\vspace{-0.3cm}
\end{table*}

\paragraph{ Action Classification on UCF and HMDB}
In \Table{ucf_hmdb} we show the top-1 accuracy on UCF and HMDB for both evaluation protocols after pretraining on Kinetics400 for 100 epochs (details in sec.~\ref{sec:impl_details}). SCVRL outperforms its corresponding baseline CVRL on UCF and HMDB when finetuning the full model. For the linear evaluation settings, we observe a marginal gain on UCF and a small loss on HMDB. We argue this is likely due to the semantic-heavy nature of these datasets (as opposed to motion, like Diving-48 and SSv2). Our fully trained model achieves inferior performance when comparing to CVRL trained using a ResNet R3D-50 architecture. Note, this gap is not caused by the method, instead, it is due to the architecture choice.

\paragraph{ Action Classification on SSv2}
In \Table{ssv2} we compare \ours{} and the baseline CVRL, both pretrained on Kinetics-400. \ours{} consistently outperforms its baseline, by $1.5\%$ on full-finetuning and $2.4\%$ on linear. 
Finally, note how we omit the recent state-of-the-art CORP~\cite{Hu_2021_ICCV_contrast_order} from the table since it is not compatible on SSv2 (i.e., differently from previous methods, it directly pretrains on the \textit{target dataset}).

\subsection{Low-Shot Learning on SSv2}
In \Table{lowshot} we evaluate our model in the context of low-shot (or semi-supervised) learning, i.e. given only a fraction of the available train-set during finetuning. Both CVRL and \ours{} are pretrained on the full Kinetics400 training set. The train-data ratio (top of the table) only applies to SSv2 finetuning. We perform the finetuning for 1\%, 5\%, 10\%, 25\%, 50\% and 100\% of SSv2 train data and show that SCVRL consistently outperforms CVRL by a significant margin. In particular, the relative gain of \ours{} is higher when the amount of supervised data for the target domain (i.e., SSv2) is lower. This shows that learning representations that capture both semantic and motion information, as in \ours{}, leads to representations that are more generalize and more transferable.

\subsection{Ablation studies}
\label{sec:ablation}

We now provide detailed ablation studies on different components and design choices. For these, we pretrain on the Kinetics-400-mini and finetune on UCF and SSv2.

\paragraph{Targeted sampling (\Table{ablation-sampling})} We evaluate the effect of our probabilistic targeted sampling on the final performance. We observe a performance boost with targeted sampling (temperature $\beta=5$) compared to uniform sampling (temperature $\beta = \infty$). Interestingly, with a low temperature value (3), which corresponds to a low entropy of the distribution of weights $w_i$, we experience a significant loss in performance. This is likely caused by the fact that this choice excludes a lot of valuable training clips and mostly focuses on a (relatively) fixed subset.

\paragraph{Head configurations (\Table{ablation-heads})} In Sec.~\ref{sec:framework} we conjectured about the importance in \ours{} of having two MLP heads, one dedicated to each contrastive objective. Here we now evaluate this choice and compare against \ours{} trained using a single head ($\phi_v=\phi_t$). The results validate our design, as they show that the performance degrades considerably when we use a single head, to the point where it is even worse than the CVRL baseline. 


\paragraph{Importance of contrastive formulation (\Table{ablation-components})} In this ablation we compare our proposed objective $\mathcal{L}^T$ against a traditional pretext task approach~\cite{noroozi-iccv2017, brattoli-cvpr2017} based on a classifier trained using a cross-entropy loss to detect if a sequence is shuffled. This loss objective is combined with the visual contrastive objective $\mathcal{L}^V$ as in \ours{}. The results show that the proposed objective achieves much higher performance for both UCF and SSv2, validating the importance of reformulating pretext tasks using contrastive learning.

\paragraph{Used output feature representation (\Table{ablation-representation})} 
As studied in the supervised setting~\cite{raghu-transformer2021}, the design choice between using a CLS token or a representation computed from spatial average pooling induces different model behaviors, especially on how localized each tokens are. 
However, it is not clear how that affects representations in a self-supervised setting.  Hence, we compare \ours{}, which is trained on the output CLS token from MViT-B, against a different version trained using the average pooled token over the time and space dimensions (AVG). We obtain stronger results for the model which operates on the CLS token.


\paragraph{Shortcuts in projection layer (\Table{ablation-kernel})} In Sec.~\ref{sec:impl_details} we discussed how for \ours{} we modify the temporal kernel size of MViT to avoid the network from learning shortcuts. We now ablate two potential choices: 2 and 3. 
The results show that SCVRL only significantly improves upon CVRL only for $k_t=2$. This is likely caused by the projection layer having directly access to the gaps between shuffled frames and with that, can detect if a sequence is shuffled. Thus, the remaining part of the transformer is not challenged by the loss. That is why it is crucial for our framework to be trained with a kernel size which is aligned with the number of frames shuffled in each group.



\section{Conclusion}
We presented \ours{}, a novel shuffled contrastive learning framework to learn self-supervised video representations that are both motion and semantic-aware. We reformulated the previously used shuffle detection pretext task in a contrastive fashion and combined it with a standard visual contrastive objective. We validated that our method better captures temporal information compared to CVRL which led to improved performances on various action recognition benchmarks.



\clearpage
\clearpage
{
    \small
    \bibliographystyle{ieee_fullname}
    \bibliography{workshop}	
}



\end{document}